# A Neuro-Fuzzy Multi Swarm FastSLAM Framework

R. Havangi, M.Teshnehlab and M.A. Nekoui

**Abstract**— FastSLAM is a framework for simultaneous localization using a Rao-Blackwellized particle filter. In FastSLAM, particle filter is used for the mobile robot pose (position and orientation) estimation, and an Extended Kalman Filter (EKF) is used for the feature location's estimation. However, FastSLAM degenerates over time. This degeneracy is due to the fact that a particle set estimating the pose of the robot loses its diversity. One of the main reasons for loosing particle diversity in FastSLAM is sample impoverishment. It occurs when likelihood lies in the tail of the proposal distribution. In this case, most of particle weights are insignificant. Another problem of FastSLAM relates to the design of an extended Kalman filter for landmark position's estimation. The performance of the EKF and the quality of the estimation depends heavily on correct a priori knowledge of the process and measurement noise covariance matrices ($Q$ and $R$) that are in most applications unknown. On the other hand, an incorrect a priori knowledge of $Q$ and $R$ may seriously degrade the performance of the Kalman filter. This paper presents a Neuro-Fuzzy Multi Swarm FastSLAM Framework. In our proposed method, a Neuro-Fuzzy extended kalman filter for landmark feature estimation, and a particle filter based on particle swarm optimization are presented to overcome the impoverishment of FastSLAM. Experimental results demonstrate the effectiveness of the proposed algorithm.

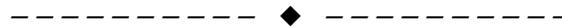

## 1 INTRODUCTION

The simultaneous localization and mapping (SLAM) is a fundamental problem of robots to perform autonomous takes such as exploration in an unknown environment. It represents an important role in the autonomy of a mobile robot. The two key computational solution to the SLAM are extended kalman filter (EKF-SLAM) and Rao-Blackwellized particle filter (FastSLAM). The EKF-SLAM approach is the oldest and the most popular approach to solve the SLAM. Until now extensive research works have been reported employing EKF to the SLAM problem [1], [2], [3]. Several applications of EKF-SLAM have been developed for indoor applications [2], [4], outdoor applications [5], underwater applications [6] and underground applications [7]. However, EKF-SLAM suffers from two major problems: the computational complexity and data association [12]. Recently, FastSLAM algorithm approach has been proposed as an alternative approach to solve the SLAM problem. FastSLAM is an instance of Rao-Blackwellized particle filter, which partitions the SLAM posterior into a localization problem and an independent landmark position estimation problem. In FastSLAM, particle filter is used for the mobile robot position estimation and EKF is used for the feature location's estimation. The key feature of FastSLAM, unlike EKF-SLAM, is the fact that data association decisions can be determined on a per-particle basis, and hence different particles can be associated with different landmarks. Each particle in FastSLAM may even have a different number of landmarks in its respective map. This characteristic gives the FastSLAM the possibility of dealing with multi-hypothesis association problem. The ability to simultaneously pursue multiple data associations makes FastSLAM significantly more robust to data association problems than algorithms based on incremental maximum likelihood data association such as EKF-SLAM. The other advantage of FastSLAM over EKF-SLAM arises from the fact that particle filters can cope with nonlinear and non-Gaussian robot motion models, whereas EKF approaches approximate such models via linear functions. However, FastSLAM also has some drawbacks. In references [13], [14], [22], [27], it has been noted that FastSLAM degenerates over time. This degeneracy is due to the fact that a particle set estimating the pose of the robot loses its diversity. One of main reasons for losing particle diversity in FastSLAM is sample impoverishment.

It occurs when likelihood lies in the tail of the proposal distribution [25]. On the other hand, FastSLAM highly relies on the number of particles to approximate the distribution density. Researchers have been trying to solve those problems in [16], [18],[25]. Another problem of FastSLAM relates to the design of EKF for landmark posi-

• R. Havangi is with the Faculty of Electrical Engineering, K.N. Toosi University of Technology,Tehran, Iran
• M.Teshnehlab S is with the Faculty of Electrical Engineering, K.N. Toosi University of Technology,Tehran , Iran
• M.A.Nekoui is with the Faculty of Electrical Engineering, K.N. Toosi University of Technology, Tehran, Iran

tion estimation. A significant difficulty in designing EKF can often be traced to incomplete a priori knowledge of the process covariance matrix $Q$ and measurement noise covariance matrix $R$. In most application these matrixes are unknown. On the other hand, an incorrect a prior knowledge of $Q$ and $R$ may seriously degrade the Kalman filter performance. In this paper to solve these problems, particle filter based on Particle Swarm Optimization (PSO) is proposed to overcome the impoverishment of FastSLAM. Also, a Neuro-Fuzzy extended kalman filter is used for landmark feature estimation.

## 2 THE SLAM PROBLEM

The goal of SLAM is to simultaneously localize a robot and determine an accurate map of the environment. To describe SLAM, let us denote the map by $\Theta$ and the pose of the robot at time by $s_t$. The map consists of a collection of features, each of which will be denoted by $\theta_n$ and the total number of stationary features will be denoted by $N$. In this situation, the SLAM problem can be formulated in a Bayesian probabilistic framework by representing each of the robot's position and map location as a probabilistic density function as:

$$p(s_t, \Theta \mid z^t, u^t, n^t) \quad (1)$$

In essence, it is necessary to estimate the posterior density of maps $\Theta$ and poses $s_t$ given that we know the observation $z^t = \{z_1,...,z_t\}$, the control input $u^t = \{u_1,...,u_t\}$ and the data association $n^t$. Here, data association represents the mapping between map points in $\Theta$ and observation in $z^t$. The SLAM problem is then achieved by applying Bayes filtering as follows:

$$p(s_t, \Theta \mid z^t, u^t, n^t) \propto p(z_t \mid s_t, \Theta, n_t)$$
$$p(s_t, \Theta \mid z^{t-1}, u^t, n^t) \quad (2)$$

with

$$p(s_t, \Theta \mid z^{t-1}, u^t, n^t) = \int p(s_t \mid s_{t-1}, u_t)$$
$$p(s_{t-1}, \Theta \mid z^{t-1}, u^t, n^t) ds_{t-1} \quad (3)$$

Where $p(s_t \mid s_{t-1}, u_t)$ is the dynamics motion model and $p(z_t \mid s_t, \Theta, n^t)$ is the measurement model. The standard Bayesian solution (2) and (3) can be extremely expensive for high dimensional map $\Theta$. However, equation (3) can not be computed in closed form. FastSLAM is an efficient algorithm for the SLAM problem that is based on a straightforward factorization as follows:

$$p(s^t, \Theta \mid z^t, u^t, n^t) = p(s^t \mid z^t, u^t, n^t)$$
$$\prod_{n=1}^{N} p(\theta_n \mid s^t, z^t, u^t, n^t) \quad (4)$$

where $s^t = \{s_1,...,s_t\}$ is a robot path or trajectory. This factorization states that the SLAM problem can be decomposed into estimating the product of a posterior over robot path and N landmark posteriors given the knowledge of the robot's path. The FastSLAM algorithm implements the path estimator $p(s^t \mid z^t, u^t, n^t)$ using a particle filter and the landmarks pose $p(\theta_n \mid s^t, z^t, u^t, n^t)$ are realized by EKF, using separate filters for different landmarks. Fig.1 shows the structure of the $M$ particles in FastSLAM. Each particle forms the following [22], [26]:

$$s_t^{[m]} = <s^{t,[m]}, \mu_{1,t}^{[m]}, \Sigma_{1,t}^{[m]},...,\mu_{N,t}^{[m]}, \Sigma_{N,t}^{[m]}> \quad (5)$$

Where $[m]$ indicates the index of the particle, and $s^{t,[m]}$ is the $m$ th particle's path estimate, and $\mu_{N,t}^{[m]}, \Sigma_{N,t}^{[m]}$ are the mean and the covariance of the Gaussian distribution representing the $n$ th feature location conditioned on the path $s^{t,[m]}$.

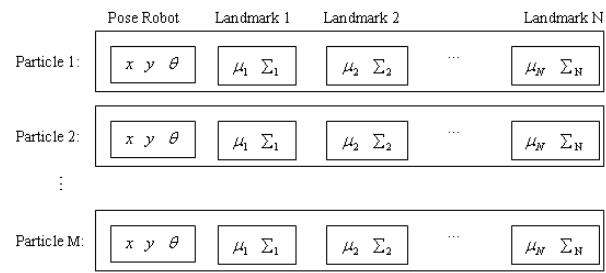

Fig.1. the structure of particles in FastSLAM

The algorithm update of posterior of FastSLAM can be described as:
1. Sample a new robot pose.
2. Update estimation of observed landmark with EKF.
3. Calculate importance weight and resample.

### 2.1 SAMPLING A NEW OF POSE

As mentioned in the previous section, FastSLAM employs a particle filter for estimating the path posterior $p(s^t \mid z^t, u^t, n^t)$ by particle filter. The particle set $S_t$ is calculated from the set $S_{t-1}$ at time $t-1$, the control $u_t$, and the observation $z_t$. In general, it is not possible to draw samples directly from the SLAM posterior. Instead, the samples are drawn from a simpler distribution called the proposal distribution $q(s^{t,[m]} \mid z^t, u^t, n^t)$. The mismatch between of SLAM posterior and proposal distribution correct using a technique called importance sampling. Therefore, in regions where the target distribution is larger than the proposal distribution, the samples are assigned a larger weight and in regions where the target distribution is smaller than the proposal distribution the samples will be given lower weights. For FastSLAM, the important weights of each particle can be calculated as:

$$w_k^i = \frac{\text{target distribution}}{\text{proposal distribution}} = \frac{p(s^{t,[m]} \mid z^t, u^t, n^t)}{q(s^{t,[m]} \mid z^t, u^t, n^t)} \quad (6)$$

As a result, the important weight of each particle is equal to the ratio of the slam posterior and the proposal distribution. The proposal $q(s^{t,[m]} \mid z^t, u^t, n^t)$ can be represented by a recursive form as:

$$q(s^{t,[m]} \mid z^t, u^t, n^t) = q(s_t^{[m]} \mid s^{t-1,[m]}, z^t, u^t, n^t)$$

$$q(s^{t-1,[m]} \mid z^t, u^t, n^t) \stackrel{Markov}{=} q(s_t^{[m]} \mid s^{t-1,[m]}, z^t, u^t, n^t) \quad (7)$$

$$q(s^{t-1,[m]} \mid z^{t-1}, u^{t-1}, n^{t-1})$$

Similarity, the posterior can also be given by a recursive form using Baye's rule as follows:

$$p(s^{t,[m]} \mid z^t, u^t, n^t) = \eta p(z_t \mid s^{t,[m]}, z^{t-1}, u^t, n^t)$$
$$p(s_t^{[m]} \mid z^{t-1}, u^t, n^t) p(s^{t-1,[m]} \mid z^{t-1}, u^{t-1}, n^{t-1}) \quad (8)$$

Where $\eta$ is a normalizing constant. By (7) and (8), a sequential importance weight of $m-th$ particle can be obtained as following:

$$w_k^i = \eta w_{k-1}^i \frac{p(z_t \mid s^{t,[m]}, z^{t-1}, u^t, n^t) p(s_t^{[m]} \mid z^{t-1}, u^t, n^t)}{q(s_t^{[m]} \mid s^{t-1,[m]}, z^t, u^t, n^t)} \quad (9)$$

The choice of the proposal distribution $q(s_t^{[m]} \mid s^{t-1,[m]}, z^t, u^t, n^t)$ is one of the most critical issues in the design of a FastSLAM. Two of those critical reasons are as follows: samples are drawn from the proposal distribution, and the proposal distribution is used to valuate important weights. The optimal importance density function that minimizes the variance of the importance weights is the following equation (FastSLAM 2.0 [9])

$$q(s^{t,[m]} \mid z^{t-1}, u^{t-1}, n^{t-1}) = p(s^{t,[m]} \mid z^{t-1}, u^t, n^{t-1}) \quad (10)$$

However, there are some special cases where the use of the optimal importance density is possible. The most popular suboptimal choice is the transitional prior (FastSLAM1.0 [11])

$$q(s^{t,[m]} \mid z^{t-1}, u^t, n^{t-1}) = p(s_t \mid u_t, s_{t-1}^{[m]}) \quad (11)$$

In this paper, FastSLAM1.0 is used due to its easy calculation. Hence, by substitution of (11) into (8), the weight's update equation is:

$$w_k^i = \eta w_{k-1}^i p(z_t \mid s^{t,[m]}, z^{t-1}, u^t, n^t) \quad (12)$$

Not that the $w_k^i$ is updated incrementally while considering a full trajectory of the robot state $s^{t,[m]}$.

## 2.2 UPDATING THE OBSERVED LANDMARK ESTIMATE

The FastSLAM represents the posterior landmark estimates $p(\theta_n \mid s^t, z^t, u^t, n^t)$ using low-dimensional EKFs. In fact FastSLAM 1.0 updates the posterior over the landmark estimates, respected by the mean $\mu_{n,t-1}^{[m]}$ and the covariance $\Sigma_{n,t-1}^{[m]}$. the updated values $\mu_{n,t}^{[m]}$ and $\Sigma_{n,t}^{[m]}$ are then added to the temporary particle set $S_t$, along with the new pose. The update depends on whether or not a landmark $n$ was observed at time $t$. For $n \neq n_t$, the posterior over the landmark remains unchanged as following [26]:

$$\mu_{n,t}^{[m]} = \mu_{n,t-1}^{[m]} \qquad \Sigma_{n,t}^{[m]} = \Sigma_{n,t-1}^{[m]} \quad (13)$$

For the observed feature $n = n_t$, the update is specified through the following equation [26]:

$$p(\theta_{n_t} \mid s^{t,(i)}, n^t, z^t) =$$
$$\frac{p(z_t \mid \theta_{n_t}, s^{t,(i)}, n^t, z^{t-1}) p(\theta_{n_t} \mid s^{t,(i)}, n^t, z^{t-1})}{p(z_t \mid s^{t,(i)}, n^t, z^{t-1})} \quad (14)$$
$$= \eta \underbrace{p(z_t \mid \theta_{n_t}, s^{t,(i)}, n^t, z^{t-1})}_{\square N(z_t, g(\theta_{n_t}, s_t^{(i)}), R_t)} \underbrace{p(\theta_{n_t} \mid s^{t,(i)}, n^t, z^{t-1})}_{\square N(\theta_{n_t}, \mu_{n,t-1}^{[i]}, \Sigma_{n,t-1}^{[i]})}$$

The probability $p(\theta_{n_t} \mid s^{t,(i)}, n^t, z^{t-1})$ at time $t-1$ is represented by a Gaussian distribution with mean $\mu_{n,t-1}^{[i]}$ and covariance $\Sigma_{n,t-1}^{[i]}$. For the new estimate at time $t$ to also be Gaussian, FastSLAM linearizes the perceptual model $p(z_t \mid \theta_{n_t}, s^{t,(i)}, n^t, z^{t-1})$ by EKF. Especially, FastSLAM approximates the measurement function $g$ by the following first degree Taylor expansion [26]:

$$g(\theta_{n_t}, s_t^{[i]}) = \underbrace{g(\mu_{n,t-1}^{[i]}, s_t^{[i]})}_{\hat{z}_t^{[i]}} + \underbrace{g'(s_t^{[i]}, \mu_{n,t-1}^{[i]})}_{G_t^{[i]}}(\theta_{n_t} - \mu_{n,t-1}^{[i]})$$
$$= \hat{z}_t^{[i]} + G_t^{[i]}(\theta_{n_t} - \mu_{n,t-1}^{[i]})$$
(15)

Under this approximation, the posterior of landmark $n_t$ is indeed Gaussian. The mean and covariance are obtained using the following measurement update:

$$\hat{z}_t = g(s_t^{[m]}, \mu_{n_t, t-1}) \quad (16)$$

$$G_{\theta_{n_t}} = \nabla_{\theta_{n_t}} g(s_t, \theta_{n_t}) \big|_{s_t = s_t^{[m]}; \theta_{n_t} = \mu_{n_t,t-1}^{[m]}} \quad (17)$$

$$Z_{n,t} = G_{\theta_{n_t}} \Sigma_{n_t,t-1}^{[m]} G_{\theta_{n_t}}^T + R_t \quad (18)$$

$$K_t = \Sigma_{n_t,t-1}^{[m]} G_{\theta_{n_t}}^T Z_{n,t}^{-1} \quad (19)$$

$$\mu_{n_t,t}^{[m]} = \mu_{n_t,t-1}^{[m]} + K_t(z_t - \hat{z}_t) \quad (20)$$

$$\Sigma_{n_t,t}^{[m]} = (I - K_t G_{\theta_{n_t}}) \Sigma_{n_t,t-1}^{[m]} \quad (21)$$

## 2.3 RESAMPLING

Sine the variance of the importance weights increase over time [23], [24], resampling plays a vital role in FastSLAM. In the resampling process, particles with low importance weight are eliminated and particles with high weights are multiplied. After, the resampling, all particle weights are then reset to

$$w_t^i = \frac{1}{N} \quad (22)$$

This enables the FastSLAM to estimate increasing environmental sates defiantly without growing a number of particles. However, resampling can delete good samples from the sample set, in the worst case, the filter diverges. The decision on how to determine the point of time of the resampling is a fundamental issue. Liu introduced the so-called effective number of particles $N_{eff}$ to estimate how well the current particle set represents the true posterior. This quality is computed as

$$N_{eff} = \frac{1}{\sum_{i=1}^{N} w^{(i)}} \quad (23)$$

Where $w^{(i)}$ refers to the normalized weight of particle $i$. The resampling process is operated whenever $N_{eff}$ is bellow a pre-defined threshold, $N_{tf}$. Here $N_{tf}$ is usually a constant value as following

$$N_{tf} = \frac{3}{4} M \quad (24)$$

Where $M$ is number of particles.

## 3 A MODIFIED FASTSLAM FRAMEWORK

In this section, a modified FastSLAM framework is presented. For this purpose, a particle filter based on PSO is presented to overcome the impoverishment of the particle filter. Also a Neuro-Fuzzy extended kalman filter (EKF) for landmark feature estimation is proposed.

### 3.1 A MODIFIED SAMPLING OF POSE

Because the proposal function is suboptimal, there are two serious problems in FastSLAM. One problem is sample impoverishment, which occurs when the likelihood $p(z_t | s^{t,[m]}, z^{t-1}, u^t, n^t)$ is very narrow or likelihood lies in the tail of the proposal distribution $p(s_t | u_t, s_{t-1}^{[m]})$. The prior distribution is effective when the observation accuracy is low. But not when prior distribution is a much broader distribution than the likelihood (such as Fig.2.). Hence, in the updating step, only a few particles will have significant importance weights.

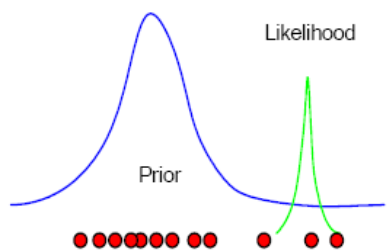

Fig.2. Prior and Likelihood

This problem implies that a large computational effort is devoted to update the particles with negligible weight. Thus, the sample set only contains few dissimilar particles and sometimes they will drop to a single sample after several iterations. As a result, important samples may be lost. Another problem of FastSLAM is the number of particles dependency that estimates the pose of the robot. If the number of particles is small, then there might not have been particles distributed around the true pose of the robot. So after several iterations, it is very difficult for particles to converge to the true pose of the robot. For standard particle filter, there is one method to solve the problem. This is to augment the number of the particles. But this would make the computational complexity unacceptable. To solve these problems of FastSLAM, particle swarm optimization is considered to optimize the sampling process of FastSLAM. In fact, PSO encourages particles to high likelihood area before the sampling process.

### 3.1.1 PARTICLE SWARM OPTIMIZATION

James Kennedy and Russell C.Eberhart [17] originally proposed the PSO algorithm for optimization. PSO is a population-based search algorithm based on the simulation of the social behavior of birds with a flock. PSO is initialized with a group of random particles and then computes the fitness of each one. Finally, it can find the best solution in the problem space via many repeating iterations. In each iteration, each particle keeps track of its coordinates which are associated with the best solution it has achieved so far (pbest) and the coordinates which are associated with the best solution achieved by any particle in the neighboring of the particle (gbest). Supposing that the search space dimension is D and number particles is N, the position and velocity of the i-th particle are represented as $x_i = (x_{i1},...,x_{iD})$ and $v_i = (v_{i1},...,v_{iD})$ respectively.

Let $P_{bi} = [p_{i1},...p_{iD}]$ denote the best position which the particle $i$ has achieved so far, and $P_g$ the best of $P_{bi}$ for any $i = 1,...,N$. The PSO algorithm could be performed by the following equations:

$$\vec{x}_i(t) = \vec{x}_i(t-1) + \vec{v}_i(t) \quad (25)$$

$$\vec{v}_i(t) = w\vec{v}_i(t-1) + c_1 r_1 (\vec{P}_{bi} - \vec{x}_i(t-1)) \\ + c_2 r_2 (\vec{P}_g - \vec{x}_i(t-1)) \quad (26)$$

Where $t$ represents the iteration number and $c_1$, $c_2$ are the learning factors. Usually $c_1 = c_2 = 2$. $r_1, r_2$ are random numbers in the interval $(0,1)$. $w$ is the inertial factor, and the bigger the value of $w$, the wider is the search range.

### 3.1.2 A PARTICLE FILTER based on PSO

As discussed in the previous section, Impoverishment occurs when the number of particles in the high likelihood area is low. In recent years, several researchers have studied this problem [16], [17], [18], [19]. The Main disadvantage of these approaches is that they are more difficult to implement than FastSLAM. Also, they are more mathematically involved. Here, we use an easy idea to solve this problem. The idea is that the sampling process particles are encouraged to be at the right place (in the region of high likelihood) by incorporating the current observation. This implies that a Simple and effective method for this purpose is the using of PSO. In fact, by using PSO, we can move all the particles towards the region of high likelihood before the sampling process. For this purpose, first we must define a fitness function. The fitness function must take the newest observations into account and also fitness function' particle with high likelihood have small value. For this purpose, we consider a fitness function as following:

$$\tilde{f}_0(x_k) = (z_t - \hat{z}_{n_t,t})^T \left[ Z_{n_t,t} \right]^{-1} (z_t - \hat{z}_{n_t,t}) \qquad (27)$$

Here, $Z_{n_t,t}$ is the residual covariance matrix defined in (12), $\hat{z}_{n_t,t}$ is the predicted measurement and $z_t$ is the actual measurement. The particles should be moved such that the fitness function is optimal. This is done by tuning the position and velocity of the PSO algorithm. The standard PSO algorithm has some parameters that need to be specified before using it. Most approaches use uniform probability distribution to generate random numbers. However it is difficult to obtain fine tuning of the solution and escape from local minima using a uniform distribution. Hence, we use velocity updates based on the Gaussian distribution. In this situation, there is no more need to specify the parameter learning factors $c_1$ and $c_2$. Furthermore, using the Gaussian PSO the inertial factor $\omega$ was set to zero and an upper bound for the maximum velocity $v_{max}$ is not necessary anymore [21]. So, the only parameter to be specified by the user is the number of particles. Initial values of particle filter are selected as the initial population of PSO. Initial velocities of PSO are set equal to zero. The PSO algorithm updates the velocity and position of each particle by following equations [21]:

$$\vec{x}_i(t) = \vec{x}_i(t-1) + \vec{v}_i(t) \qquad (28)$$

$$\vec{v}_i(t) = |randn|(P_{pbest} - \vec{x}_i(t-1)) + |randn|(P_{gbest} - \vec{x}_i(t-1)) \qquad (29)$$

PSO moves all particles towards the high likelihood regions which is the global best of PSO. When the best fitness value reaches a certain threshold, the optimized sampling process is stopped.

With this set of particles the sampling process will be done on the basis of proposal distribution. The Corresponding weights will be as follows:

$$w_i^k = p(z_t | s^{t,[m]}, z^{t-1}, u^t, n^t) \qquad (30)$$

Where

$$p(z_t | s^{t,[m]}, z^{t-1}, u^t, n^t) = \frac{1}{\sqrt{(2\pi)|Z_{n_t,t}|}} \exp \\ \{-\frac{1}{2}(z_t - \hat{z}_{n_t,t})^T \left[ Z_{n_t,t} \right]^{-1} (z_t - \hat{z}_{n_t,t})\} \qquad (31)$$

The flowchart of the proposed algorithm to pose estimation is shown in fig.3.

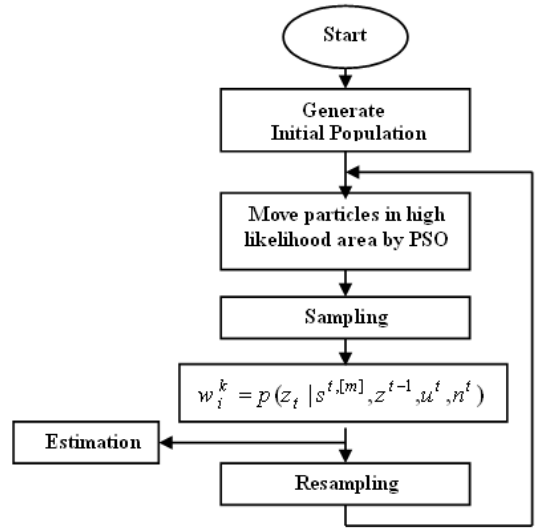

Fig3. Modified particle filter for FastSLAM (pose estimation)

As observed, this algorithm can be described with the following steps

**Step1.** General initial population
Initialization-randomly generation initial pose of mobile robot
- Initialized particle velocity
- Initial particle position
- Initialized particle fitness value
- Initialized pbest and gbest

**Step2.** Improve Sampling by PSO

**Step5.** Sampling
Sampling is obtaining samples from the prior

**Step3.** Update
On receipt of the measurement saw $z_k$, evaluate the likelihood of each prior sample and obtain a normalized-weight for each sample.

**Step4.** Resampling
Multiply Suppress samples with high /low importance weights.

**Step5.** Prediction
Each pose is passed through the system model to obtain samples from the prior

**Setp6.** Increase time $t$ and iterate from step 2 to step 4.

### 3.2 FEATURE UPDATE BY ADAPTIVE NEURO-FUZZY INFERENCE SYSTEM (ANFIS)

As stated earlier, FastSLAM represents the conditional landmark estimates $p(\theta_n | s^t, z^t, u^t, n^t)$ by EKF. The traditional EKF assumes complete a priori knowledge of the process and measurement noise statistics; matrices Q and R. However, in most applications these matrixes are unknown. An incorrect a prior knowledge of $Q$ and $R$ may lead to performance degradation [15] and it can even lead to practical divergence [16]. One of the efficient ways to overcome the above weakness is to use an adaptive algorithm. Two major approaches that have proposed for adaptive EKF are multiple model adaptive estimation (MMAE) and innovation adaptive estimation (IAE) [15]. Although the implementation of these approaches is different, they both share the same concept of utilizing new statistical information obtained from the residual (innovation) sequence. As landmarks are static, we assume that the noise covariance $Q$ is completely known. Hence, the algorithm to estimate the measurement noise covariance $R$ can be derived. The adaptation is adaptively adjusting the measurement noise covariance matrix R by using Neuro-fuzzy System. In this case, an innovation based adaptive estimation (IAE) algorithm to adapt the measurement noise covariance matrix R is derived. The technique known as covariance-matching is used. The basic idea behind this technique is to make the actual value of the covariance of the residual to be consistent with its theoretical value. The innovation sequence $r_k = (z_t - \hat{z}_{n_t,t})$ has a theoretical covariance that is obtained from the EKF algorithm

$$S_k = G_{\theta_{n_t}} \Sigma^{[m]}_{n_t,t-1} G^T_{\theta_{n_t}} + R_t \qquad (32)$$

The actual residual covariance $\hat{C}_k$ can be approximated by its sample covariance, through averaging inside a moving window of size N as following

$$\hat{C}_k = \frac{1}{N} \sum_{i=k-N+1}^{k} (r_i^T r_i) \qquad (33)$$

Where $i_0$ is first sample inside the estimation window. If the actual value of covariance $\hat{C}_k$ has discrepancies with its theoretical value, then the diagonal elements of $R_k$ based on the size of this discrepancy can be adjusted. The objective of these adjustments is to correct this mismatch as far as possible. The size of the mentioned discrepancy is given by a variable called the degree of mismatch ($DOM_k$), defined as

$$DOM_k = S_k - \hat{C}_k \qquad (34)$$

The basic idea used to adapt the matrix $R_k$ is as follows: from equation (24) an increment in $R_k$ will increase $S_k$ and vice versa. Thus, $R_k$ can be used to vary $S_k$ in accordance with the value of $DOM_k$ in order to reduce the discrepancies between $S_k$ and $\hat{C}_k$. The adaptation of the $(i,i)$ th element of $R_k$ is made in accordance with the $(i,i)$ th element of $DOM_k$. The general rules of adaptation are as follows:

If $DOM_k(i,i) \cong 0$ then maintain $S_k$ unchanged

If $DOM_k(i,i) > 0$ then decrease $S_k$

If $DOM_k(i,i) \cong 0$ then increase $S_k$

In this paper the IAE adaptive scheme of the EKF coupled with adaptive Neuro-fuzzy inference system (ANFIS) is presented to adjust $R$. As the size $DOM_k$ and $R$ is two, two system ANFIS to adjust EKF is used. The structure of one of these systems is described in the next section.

### 3.2.1 THE ANFIS ARCHITECTURE

The ANFIS model has been considered as a two-input-single-output system. The inputs of ANFIS are $DOM_k$ and $DeltaDOM_k$. Here, $DeltaDOM_k$ is defined as following

$$DeltaDOM_k = DOM_k - DOM_{k-1} \qquad (35)$$

Fig.4 and Fig.5 presents membership functions for $DOM_k(i,i)$ and $DeltaDOM_k$.

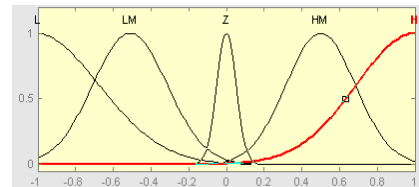
Fig.4. Membership function $DOM_k$

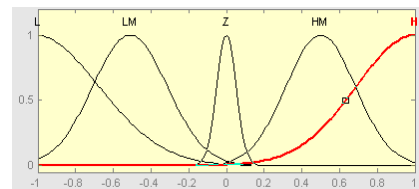
Fig .5.Membership function $DeltaDOM_k$

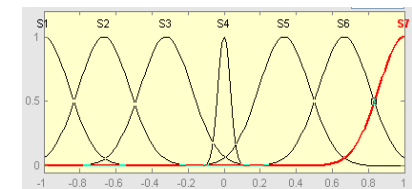
Fig .6.Membership function $AjdR_k$

Finally, adjustments of $R_k$ is performed using the following relation

$$R_k = R_k + \Delta R_k \qquad (29)$$

where $\Delta R_k$ is the ANFIS output and membership function of $\Delta R_k$ is shown in Fig.6. This ANFIS is a five layers network as shown in Fig.7.

Let $u_i^l$ and $o_i^l$ denote the input to output from the $i$ th node of the $l$ th layer, respectively. To provide a clear understanding of an ANFIS, the function of layer 1 to layer 5 are defined as follows:

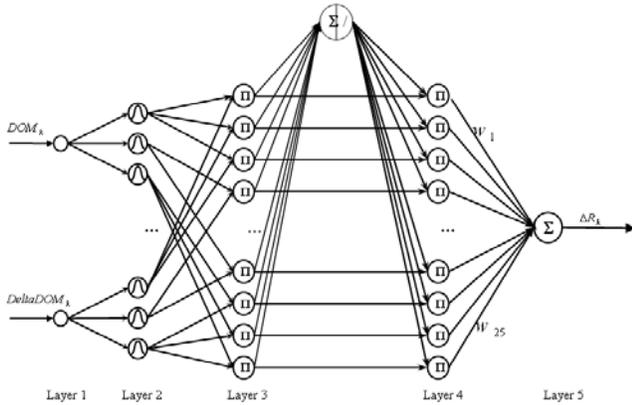

Fig .7. The Neuro-fuzzy system for feature update

Table 1
Rule Table

| DeltaDOM$_k$ / DOM$_k$ | L | LM | Z | HM | H |
|---|---|---|---|---|---|
| L | S7 | S7 | S6 | S5 | S4 |
| LM | S7 | S6 | S5 | S4 | S3 |
| Z | S6 | S5 | S4 | S3 | S2 |
| HM | S5 | S4 | S3 | S2 | S1 |
| H | S4 | S3 | S2 | S1 | S1 |

Layer 1: The node in this layer only transmits input values to the next layer directly, i.e.

$$o_i^1 = u_i^1 \quad (36)$$

Layer2: In this layer, each node only performs a membership function. Here, the input variable is fuzzified employing five membership functions (MFs). The output of the $i$ th MF is given as:

$$o_{ij}^2 = \mu_{ij}(u^2) = \exp\left\{-\frac{(u_{ij}^2 - m_{ij})^2}{(\delta_{ij})^2}\right\} \quad (37)$$

Where $m_{ij}$ and $\delta_{ij}$, respectively, are the mean and with of the Gaussian membership function. The subscript $ij$ indicates the $j$th term of the $i$th input. Each node in this layer has two adjustable parameters: $m_{ij}$ and $\delta_{ij}$

Layer3: The nodes in this layer are rule nodes. The rule node performs a fuzzy and operation (or product inference) for calculating the firing strength.

$$o_l^3 = \prod_i u_i^3 \quad (38)$$

Layer4: The node in this layer performs the normalization of firing strengths from layer 3,

$$o_l^4 = \frac{u_l^4}{\sum_{l=1}^{9} u_l^4} \quad (39)$$

Layer5: This layer is the output layer. The link weights in this layer represent the singleton constituents ($W_i$) of the output variable. The output node integrates all the normalization firing strength from layer 4 with the corresponding singleton constituents and acts as defuzzfier,

$$\Delta R_i = \sum_{l=1}^{25} u_l^5 w_l \quad (40)$$

The fuzzy rules which complete the ANFIS rule base as table1.

### 3.2.2 LEARNING ALGORITHM

The aim of the training algorithm is to adjust the network weights through the minimization of the following cast function:

$$E = \frac{1}{2} e_k^2 \quad (41)$$

Where

$$e_k = S_k - \hat{C}_k \quad (42)$$

By using the bake propagation (BP) learning algorithm, the weighting vector of ANFIS is adjusted such that the error defined in (41) is less of than a desired threshold value after a given number of training cycles. The well-known BP algorithm may be written as:

$$W(k+1) = W(k) + \eta\left(-\frac{\partial E(k)}{\partial W(k)}\right) \quad (43)$$

Here $\eta$ and $W$ represent the learning rate and tuning parameter of ANFIS respectively. Let $W = [m, \sigma, w]^T$ be the weighting vector of ANFIS. The gradient of $E$ with respect to an arbitrary weighting vector $W$ as following:

$$\frac{\partial E(k)}{\partial W(k)} = -e(k)\left(\frac{\partial \Delta R(k)}{\partial W(k)}\right) \quad (44)$$

By the recursive application of the chain rule, the error term for each layer is first calculated, and then the parameters in the corresponding layers are adjusted.

## 4 IMPLEMENTATION AND RESULTS

Simulation experiments have been carried out to evaluate the performance of the proposed approach in comparison with the classical method. The proposed solution for the SLAM problem has been tested for the benchmark environment, with varied number and position of the landmarks, available in [20] .Fig.8 shows the robot trajectory and landmark location. The star points (*) depict location of the landmarks that are known and stationary in the environment. The state of the robot can be modeled as $(x, y, \theta)$ that $(x, y)$ are the Cartesian coordinates and $\theta$ is the orientation respectively to the global environment. The kinematics equations for the mobile robot are in the

following form

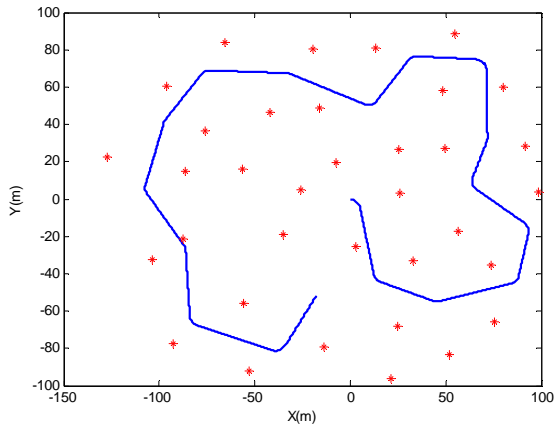

Fig.8. The experiment environment. The star point "*" denote the landmark positions and blue line is the path of robot.

$$\begin{bmatrix} \dot{x} \\ \dot{y} \\ \dot{\phi} \end{bmatrix} = \begin{bmatrix} (V + v_v)\cos(\phi + [\gamma + v_\gamma]) \\ (V + v_v)\sin(\phi + [\gamma + v_\gamma]) \\ \dfrac{(V + v_v)}{B}\sin(\gamma + v_\gamma) \end{bmatrix} \quad (45)$$

Where $B$ is the base line of the vehicle and $u = \begin{bmatrix} V & \gamma \end{bmatrix}^T$ is the control input at time $t$ consisting of a velocity input $V$ and a steer input $\gamma$. The process noise $v = \begin{bmatrix} v_v & v_\gamma \end{bmatrix}^T$ is assumed to be Gaussian. The vehicle is assumed to be equipped with a range-bearing sensor that provides a measurement of range $r_i$ and bearing $\theta_i$ to an observed feature $\rho_i$ relative to the vehicle. The observation $z$ of feature $\rho_i$ in the map can be expressed as:

$$\begin{bmatrix} r_i \\ \theta_i \end{bmatrix} = \begin{bmatrix} \sqrt{(x - x_i)^2 + (y - y_i)^2} + \omega_r \\ \tan^{-1}\dfrac{y - y_i}{x - x_i} - \phi + \omega_\theta \end{bmatrix} \quad (46)$$

Where $(x_i, y_i)$ is the landmark position in map and $W = \begin{bmatrix} \omega_r & \omega_\theta \end{bmatrix}^T$ related to observation noise.

The initial position of the robot is assumed to be $x_0 = 0$. The robot moves at a speed 3m/s and with a maximum steering angle 30 deg. Also, the robot has 4 meters wheel base and is equipped with a range-bearing sensor with a maximum range of 20 meters and a 180 degrees frontal field-of-view. The control noise is $\sigma_v = 0.3$ m/s and $\sigma_\gamma = 3^o$. A control frequency is 40 HZ and observation scans are obtained at 5 HZ. The measurement noise is 0.2 m in range and $1^o$ in bearing. Data association is assumed known. At first, the performance of the two algorithms can be compared by keeping the noises level (process noise and measurement noise) and varying the number of particles. Fig.9 to Fig.16 shows the performance of the two algorithms for number of particles equivalent to 20 and 5. As observed, modified FastSLAM is more accurate than the FastSLAM. Also, performance of modified FastSLAM does not depend on the number of particles while the performance of FastSLAM highly depends on the number of particles. For very low numbers of particles FastSLAM diverges while modified FastSLAM is completely robust. This is because PSO in the modified FastSLAM places the particles in the high likelihood region. In addition, we observed that the modified FastSLAM requires fewer particles than FastSLAM in order to achieve a given level of accuracy for state estimates.

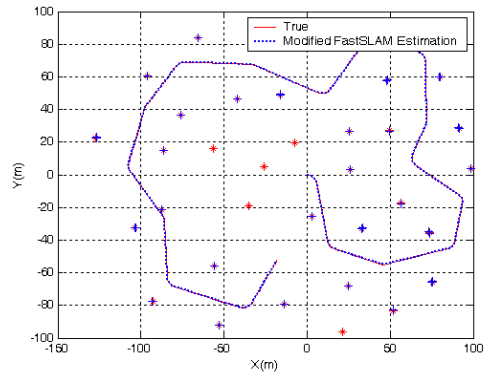

Fig.9. Estimated robot path and estimated landmark using modified FastSLAM with true robot path and true landmark. The "…" is path estimated, the "+" are the estimated landmark positions. In this experiment the control noise is $\sigma_v = 0.3$ m/s, $\sigma_\gamma = 3^o$. The measurement noise is $\sigma_r = 0.2$ m, $\sigma_\theta = 1$ deg and number of particles is 20 (n=20).Also, the results obtain over 50 Monte Carlo runs.

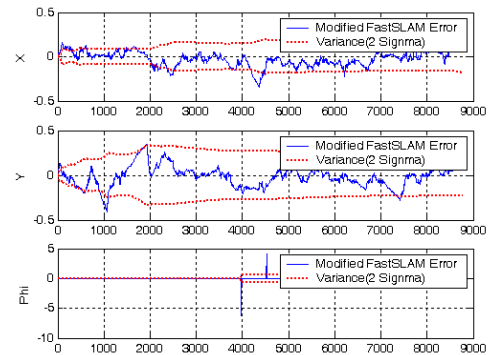

Fig.10.Estimated pose error with $2 - \sigma$ bound using modified FastSLAM .In this experiment the control noise is $\sigma_v = 0.3$ m/s, $\sigma_\gamma = 3^o$. The measurement noise is $\sigma_r = 0.2$ m, $\sigma_\theta = 1$ deg and number of particles is 20 (n=20).Also, the results obtain over 50 Monte Carlo runs.

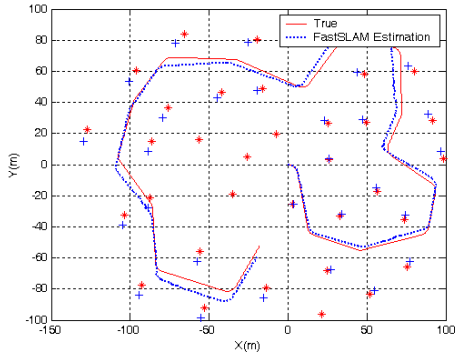

Fig.11. Estimated robot path and estimated landmark using FastSLAM with true robot path and true landmark. The "…" is path estimated, the "+" are the estimated landmark positions. In this experiment the control noise is $\sigma_v = 0.3 \text{ m/s}$, $\sigma_\gamma = 3^o$. The measurement noise is $\sigma_r = 0.2 \text{ m}$, $\sigma_\theta = 1 \deg$ and number of particles is 20 (n=20). Also, the results obtain over 50 Monte Carlo runs.

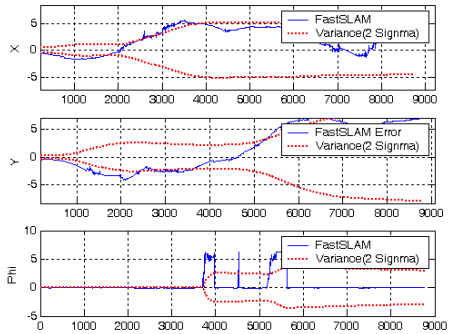

Fig.12. Estimated pose error with $2-\sigma$ bound using modified FastSLAM. In this experiment the control noise is $\sigma_v = 0.3 \text{ m/s}$, $\sigma_\gamma = 3^o$. The measurement noise is $\sigma_r = 0.2 \text{ m}$, $\sigma_\theta = 1 \deg$ and number of particles is 20 (n=20). Also, the results obtain over 50 Monte Carlo runs.

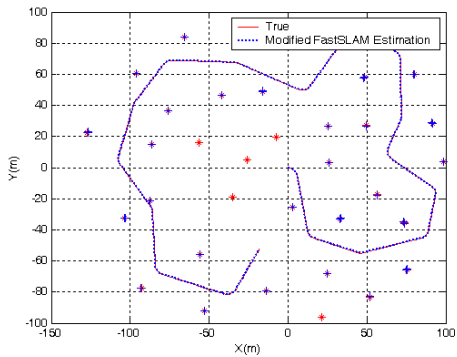

Fig.13. Estimated robot path and estimated landmark using FastSLAM with true robot path and true landmark. The "…" is path estimated, the "+" are the estimated landmark positions. In this experiment the control noise is $\sigma_v = 0.3 \text{ m/s}$, $\sigma_\gamma = 3^o$. The measurement noise is $\sigma_r = 0.2 \text{ m}$, $\sigma_\theta = 1 \deg$ and number of particles is 5 (n=5). Also, the results obtain over 50 Monte Carlo runs.

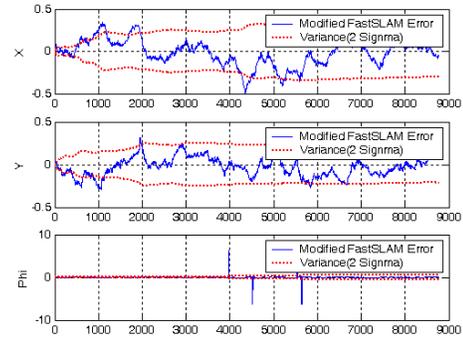

Fig.14. Estimated pose error with $2-\sigma$ bound using modified FastSLAM. In this experiment the control noise is $\sigma_v = 0.3 \text{ m/s}$, $\sigma_\gamma = 3^o$. The measurement noise is $\sigma_r = 0.2 \text{ m}$, $\sigma_\theta = 1 \deg$ and number of particles is 5 (n=5). Also, the results obtain over 50 Monte Carlo runs.

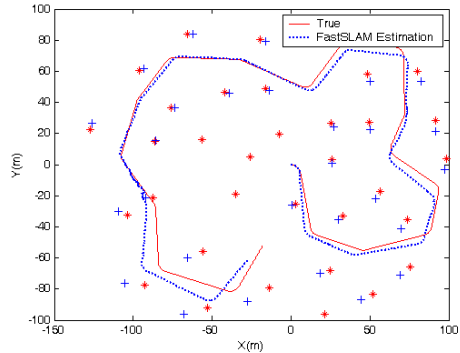

Fig.15. Estimated robot path and estimated landmark using FastSLAM with true robot path and true landmark. The "…" is path estimated, the "+" are the estimated landmark positions. In this experiment the control noise is $\sigma_v = 0.3 \text{ m/s}$, $\sigma_\gamma = 3^o$. The measurement noise is $\sigma_r = 0.2 \text{ m}$, $\sigma_\theta = 1 \deg$ and number of particles is 5 (n=5). Also, the results obtain over 50 Monte Carlo runs.

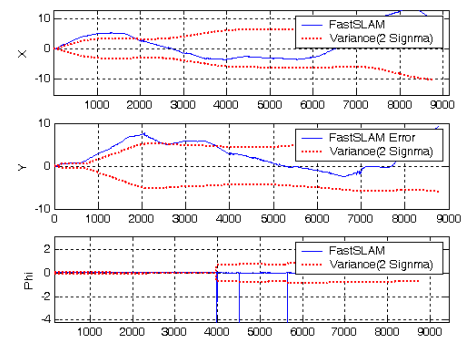

Fig.16. Estimated pose error with $2-\sigma$ bound using modified FastSLAM. In this experiment the control noise is $\sigma_v = 0.3 \text{ m/s}$, $\sigma_\gamma = 3^o$. The measurement noise is $\sigma_r = 0.2 \text{ m}$, $\sigma_\theta = 1 \deg$ and number of particles is 5 (n=5). Also, the results obtain over 50 Monte Carlo runs.

Now, we compare the performance of the two algorithms while varying the level of measurement noise as $\sigma_r = 0.01$ a, $\sigma_\theta = 0.1$ and fixing the control noise i.e. $\sigma_v = 0.3$ m/s , $\sigma_\gamma = 3^o$. All experiments run with 5 particles in this case. From Fig.17 to Fig.20 we observe that the performance of our proposed method will outperform classical FastSLAM.

This is because sample impoverishment occurs in FastSLAM. In fact, because the proposal distribution is broader than likelihood in FastSLAM, the weight of most of the particles are insignificant when their likelihoods are evaluated. As a result, the estimation accuracy of FastSLAM will decrease whenever the robot's motion is noisier and the range-bearing sensor is more accurate. In this situation, the performance of FastSLAM is reduced mostly while the performance of the modified FastSLAM is almost fixed. For very low values of measurement error FastSLAM clearly begins to diverge while the proposed algorithm is robust.

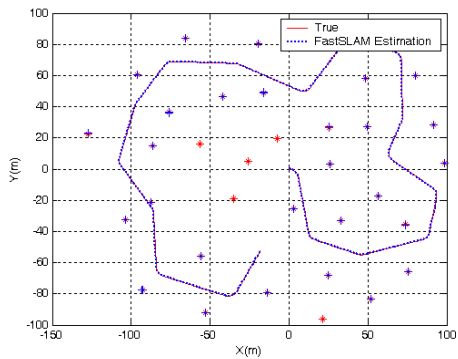

Fig.17. Estimated robot path and estimated landmark using modified FastSLAM with true robot path and true landmark. The "…" is path estimated, the "+" are the estimated landmark positions In this experiment the control noise is $\sigma_v = 0.3$ m/s , $\sigma_\gamma = 3^o$ .The measurement noise is $\sigma_r = 0.01$ m , $\sigma_\theta = 0.1$ deg and number of particles is 5 (n=5).Also, the results obtain over 50 Monte Carlo runs.

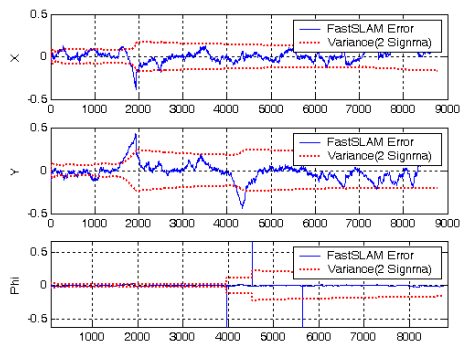

Fig.18. Estimated pose error with $2-\sigma$ bound using modified FastSLAM. In this experiment the control noise is $\sigma_v = 0.3$ m/s , $\sigma_\gamma = 3^o$ .The measurement noise is $\sigma_r = 0.01$ m , $\sigma_\theta = 0.1$ deg and number of particles is 5 (n=5).Also, the results obtain over 50 Monte Carlo runs.

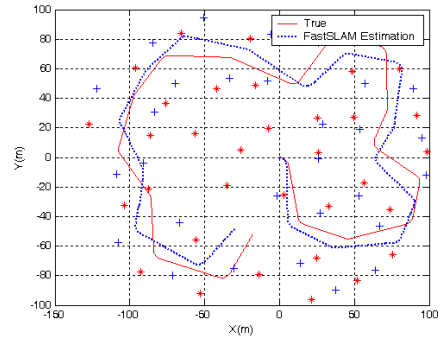

Fig.19. Estimated robot path and estimated landmark using FastSLAM with true robot path and true landmark. The "…" is path estimated, the "+" are the estimated landmark positions. In this experiment the control noise is $\sigma_v = 0.3$ m/s , $\sigma_\gamma = 3^o$ . The measurement noise is $\sigma_r = 0.01$ m , $\sigma_\theta = 0.1$ deg and number of particles is 5 (n=5).Also, the results obtain over 50 Monte Carlo runs.

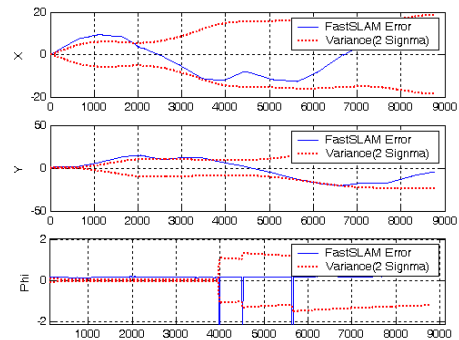

Fig.20. Estimated pose error with $2-\sigma$ bound using modified FastSLAM. In this experiment the control noise is $\sigma_v = 0.3$ m/s , $\sigma_\gamma = 3^o$ . The measurement noise is $\sigma_r = 0.01$ m , $\sigma_\theta = 0.1$ deg and number of particles is 5 (n=5).Also, the results obtain over 50 Monte Carlo runs.

Finally, we study affect of the designed ANFIS on the performance of FastSLAM. For this purpose, we consider the situation where measurement noise is wrongly considered as $\sigma_r = 0.5$ , $\sigma_\theta = 3.0$ and control noise is rightly considered as $\sigma_v = 0.3$ m/s, $\sigma_\gamma = 3^o$ .The performance of the modified FastSLAM is compared with classical FastSLAM that its measurement covariance matrix R is kept static throughout the experiment. The proposed algorithm starts with a wrongly known statistics and then adapts the $R_k$ matrix through ANFIS and attempts to minimize the mismatch between the theoretical and actual values of the innovation sequence. The free parameters of ANFIS are automatically learned by SD during training. Fig.21-22 and Fig.24-25 show the comparison between the proposed algorithm and the FastSLAM when the particle number is 20. It can be seen clearly that the results of the proposed algorithm are better than that of classical FastSLAM. This is because modified FastSLAM adaptively tuned the measurement covariance matrix R while covariance matrix measurement R in standard FastSLAM is kept fixed over time as shown in Fig.23 and Fig.26. Also Fig.23 and Fig.26 show that measurement covariance matrix measurement R converges to the actual covariance matrix R in our proposed method.

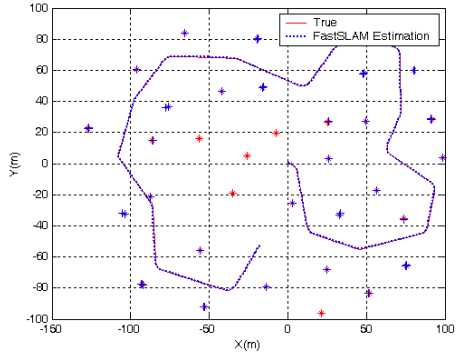

Fig.21. Estimated robot path and estimated landmark using modified FastSLAM with true robot path and true landmark. The "…" is path estimated, the "+" are the estimated landmark positions. In this experiment, measurement noise is wrongly considered as $\sigma_r = 0.5$, $\sigma_\theta = 3.0$ and control noise is truly considered as $\sigma_v = 0.3$ m/s, $\sigma_\gamma = 3^o$. Also, number of particles is 5 (n=5) and the results obtain over 50 Monte Carlo runs.

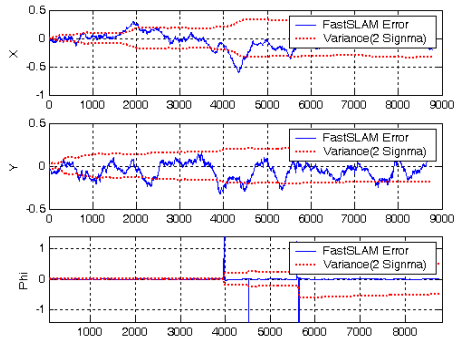

Fig.22. Estimated pose error with $2-\sigma$ bound using modified FastSLAM. In this experiment, measurement noise is wrongly considered as $\sigma_r = 0.5$, $\sigma_\theta = 3.0$ and control noise is truly considered as $\sigma_v = 0.3$ m/s, $\sigma_\gamma = 3^o$. Also, number of particles is 5 (n=5) and the results obtain over 50 Monte Carlo runs.

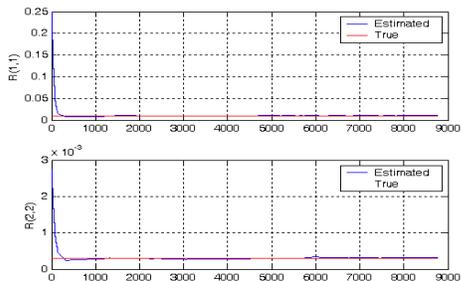

Fig.23. A modified FastSLAM (n=5)

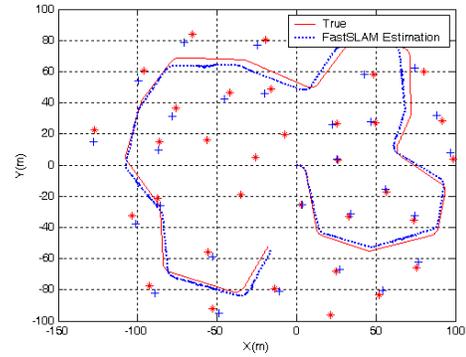

Fig.24. Estimated robot path and estimated landmark using FastSLAM with true robot path and true landmark. The "…" is path estimated, the "+" are the estimated landmark positions. In this experiment, measurement noise is wrongly considered as $\sigma_r = 0.5$, $\sigma_\theta = 3.0$ and control noise is truly considered as $\sigma_v = 0.3$ m/s, $\sigma_\gamma = 3^o$. Also, number of particles is 5 (n=5) and the results obtain over 50 Monte Carlo runs.

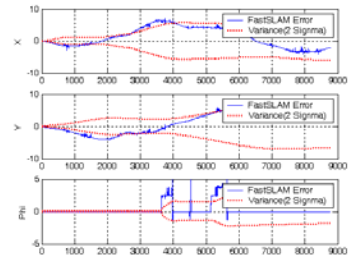

Fig.25. Estimated pose error with $2-\sigma$ bound using FastSLAM. In this experiment, measurement noise is wrongly considered as $\sigma_r = 0.5$, $\sigma_\theta = 3.0$ and control noise is truly considered as $\sigma_v = 0.3$ m/s, $\sigma_\gamma = 3^o$. Also, number of particles is 5 (n=5) and the results obtain over 50 Monte Carlo runs.

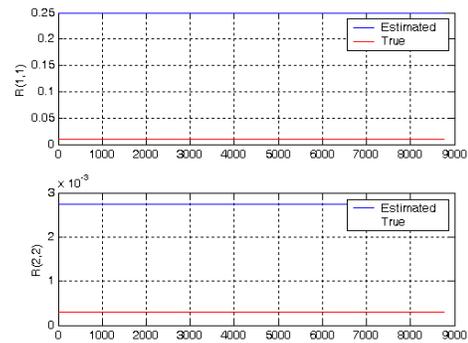

Fig.26. A FastSLAM (n=5)

## 5 CONCLUSION

In this Paper, two problems of FastSLAM are improved. The first problem is that FastSLAM degenerates over time due to the loss of particle diversity. One of the main reasons for losing particle diversity is sample impoverishment. It occurs when likelihood lies in the tail of the proposal distribution. In this case, most of particles weights are insignificant. Another problem of FastSLAM relates to the design of EKF for the landmark position's estimation. A significant difficulty in designing the extended kalman filter can often be traced to incomplete a priori knowledge of the process covariance matrix $Q$ and the measurement noise covariance matrix $R$. On the other hand, an incorrect a prior knowledge of $Q$ and $R$ may seriously degrade the Kalman filter performance. This paper presents a modified FastSLAM framework by soft computing. In the proposed method, a Neuro-Fuzzy extended kalman filter for landmark feature estimation, and a particle filter based on particle swarm optimization is presented to overcome the impoverishment of FastSLAM. Finally, Experimental results confirm the effective of the proposed algorithm. The main advantage of our proposed method is its more consistency than classical FastSLAM. This is because in our proposed method, the theoretical value of the innovation sequence matches with its real value. Also, when motion model is noisier than measurement, the performance of the proposed method outperforms the standard method.

## REFERENCES

[1] M. W. M. G. Dissanayake, P. Newman, S. Clark, and H. F. Durrant Whyte, "A solution to the simultaneous localization and map building (SLAM) problem", IEEE Tran Robot Automat, vol. 17, no. 3, pp.229–241, Jun. 2001.

[2] S. Thrun, D. Fox, and W. Burgard, "A probabilistic approach to concurrent mapping and localization for mobile robots," Mach. Learn., vol. 31, pp. 29–53, 1998.

[3] R. Smith and P. Cheeseman, "On the representation and estimation of spatial uncertainty," Int. J. Robot. Res., vol. 5, no. 4, pp. 56–68, 1986.

[4] M. Bosse, J. Leonard, and S. Teller, J. Leonard, J. D. Tard'os, S. Thrun, and H. Choset, Eds., "Large-scale CML using a network of multiple local maps," in workshop Notes of the ICRA Workshop on Concurrent Mapping and Localization for Autonomous Mobile Robots(W4),Washington, D.C., 2002.

[5] T. Bailey, "Mobile robot localization and mapping in extensive outdoor environments," Ph.D. dissertation, Univ. Sydney, Sydney, NSW, Australia, 2002.

[6] S. Williams, G.Dissanayake, and H. F. Durrant-Whyte, "Towards terrain- aided navigation for underwater robotics," Adv. Robot., vol. 15,no. 5, pp. 533–550, 2001.

[7] S. Thrun, D. H¨ahnel, D. Ferguson, M. Montemerlo, R. Triebel, W.Burgard, C. Baker, Z. Omohundro, S. Thayer, and W. Whittaker, "A system for volumetric robotic mapping of abandoned mines," in Proc. IEEE Int. Conf Robot. Automat. Taipei, Taiwan, May 2003, pp.4270–4275.

[8] A. Eliazar, R. Parr," DP-SLAM: fast, robust simultaneous localiza tion and mapping without predetermined landmarks", in: Proceedings of the International Joint Conference on Artificial Intelligence, 2003.

[9] D. Roller, M. Montemerlo, S.Thrun, B. Wegbreit, "Fastslam 2.0: an improved particle filtering algorithm for simultaneous localization and mapping that provably converges", in: Proceedings of the International Joint Conference on Artificial Intelligence, 2003.

[10] K. Murphy, "Bayesian map learning in dynamic environments", in: Proc. of the Conf. on Neural Information Processing Systems (NIPS), Denver, CO, USA, 1999, pp.1015–1021.

[11] Montemerlo M, Thrun S, Koller D, Wegbreit B (2002), " Fast SLAM: a factored solution to the simultaneous localization and mapping problem", In: Proc. AAAI national conf. artif. intell., Edmonton,AB, Canada.

[12] S. Thrun, W. Burgard, and D. Fox, "Probabilistic robotics" , MIT Press, Cambridge, 2005.

[13] T. Bailey, J. Nieto, and E. Nebot, "Consistency of the FastSLAM algorithm", IEEE International Conference on Robotics and Automation.pp.424-429, 2006.

[14] M. Boli´c, P. M. Djuri´c, and S. Hong,"Resampling algorithms for particle filters: a computational complexity perspective" ,Journal on Applied Signal Processing, no. 15, pp. 2267-2277, 2004.

[15] R. K. Mehra, "On the identification of variances and adaptive Kalman filtering", IEEE Trans. Autom. Control, vol. AC-15, no. 2, pp. 175–184, Apr. 1970.

[16] C. Kim, R. Sakthivel and W. K. Chung, "Unscented FastSLAM: A Robust and Efficient Solution to the SLAM Problem", IEEE Trans. Robot., vol. 24, no. 4, 2008.

[17] M. Pitt and N. Shephard, "Filtering via simulation: Auxiliary particle filters," J. Amer. Statist. Assoc., vol. 94, no. 446, pp. 590–599, 1999.

[18] C. Kim, R. Sakthivel, andW. K. Chung, "Unscented FastSLAM: A Robust algorithm for the simultaneous localization and mapping problem", in Proc. IEEE Int. Conf. Robot. Autom., 2007, pp. 2439–2445.

[19] X. Wang and H. Zhang,"A UPF-UKF framework for SLAM", IEEE Intl. Conf. on Robotics and Automation, 2007, pp.1664-1669.

[20] Available:http://www.personal.acfr.usyd.edu.au/tbailey/softw are/index .html, (2008, Jun.).

[21] R. A. Krohling," Gaussian swarm: a novel particle swarm optimization algorithm", In Proceedings of the IEEE Conference on Cybernetics and Intelligent Systems (CIS), Singapore, pp.372-376, 2004.

[22] M.Montemerlo," FastSLAM: A Factored Solution to the Simultaneous Localization and Mapping Problem With Unknown Data Association",PhD thesis, Carnegie Mellon University, 2003.

[23] G. Grisetti, C. Stachniss and W. Burgard, "Improved Techniques for Grid Mapping with Rao-Blackwellized Particle Filters," IEEE Trans. on Robotics, vol. 23(1), 2007, pp 34-46.

[24] G. Grisetti, C. Stachniss and W. Burgard, "Improving Grid-based SLAM with Rao-Blackwellized Particle Filters by Adaptive Proposal and Selective Resampling", IEEE Intl. Conf. on Robotics and Automation, 2005, pp 2443-2448.

[25] N. Kwak, I. K. Kim, H. C. Lee and B. H. Lee, "Adaptive Prior Boosting Technique for the Efficeint Sample Size in FastSLAM", Proceedings of the IEEE/RSJ International Conference on Intelligent Robots and Systems (San Diego, CA,2007) pp. 630–635.

[26] Thrun, S., Montemerlo, M., Koller, D., Wegbreit, B., Nieto, J., Nebot,E.," FastSLAM: An efficient solution to the simultaneous Localization and mapping problem with unknown data association", Journal of Machine Learning Research (2004).

[27] Liang Zhang, Xu-jiong Meng, Yao-wu Chen," onvergence and Consistency analysis for FastSLAM", 2009 IEEE.